# An Agentic Workflow for Legacy HPC Modernization: Converting the Two-Electron-Integral Core of GAMESS

Yuzhong Shen
*Dept. of Electrical and Computer Engineering*
*Old Dominion University*
Norfolk, VA, USA
yshen@odu.edu

Masha Sosonkina
*Dept. of Electrical and Computer Engineering*
*Old Dominion University*
Norfolk, VA, USA
msosonki@odu.edu

Peng Xu
*Dept. of Chemistry*
*Iowa State University*
*and Ames National Laboratory*
Ames, IA, USA
pxu@iastate.edu

Mark S. Gordon
*Dept. of Chemistry*
*Iowa State University*
*and Ames National Laboratory*
Ames, IA, USA
mgordon@iastate.edu

***Abstract*—Modernizing legacy Fortran is a problem of volume: the transformations are individually routine, but the codebases can be enormous, and across much of computational science the work simply goes undone. We propose an agentic workflow that takes this work on at production scale, and we set out to measure how far such delegation can reach. In this work, three prompt-specialized agent roles operate under a version-controlled specification that the agents themselves authored and revised, while humans hold a small number of gates. The arrangement is kept safe by an exact verification oracle inherited from the domain, and the boundary of safe delegation lies exactly where that oracle stops seeing.**

**We apply the proposed workflow in a case study, converting the two-electron-integral routines of GAMESS (General Atomic and Molecular Electronic Structure System), a mature quantum-chemistry package with a 48-year development history, from fixed-form Fortran 77 to free-form Fortran 2008. The scope of this work was twelve source files, 56,448 lines, and 225 subroutines for computing electron repulsion integrals. The agents ran as three Claude Code roles in isolated worktrees, and the work spanned four Claude model generations. Because the GAMESS group ships a standard test suite whose printed energies its user community treats as canonical, we could adopt bit-for-bit reproduction of those energies as the merge criterion, where a deviation in the twelfth decimal place counts as a failure rather than drift. All twelve source files pass a 51-test validation battery comprising the 49 standard GAMESS tests and two additional calculations, and across 612 test runs the number of chemistry-relevant differences is zero, and every file also passes the Jenkins tests that are used for continuous integration.**

## I. Introduction

Fortran has been the working language of computational science since its introduction in the 1950s, and it remains so today. Climate and weather models, computational chemistry and materials codes, and the dense linear-algebra libraries beneath much of scientific computing are still written and actively developed in Fortra [1, 2]. The language has grown across a long line of standards, from FORTRAN 66 and 77 through Fortran 90, 95, 2003, 2008, and 2018, each adding what the previous generation lacked: free-form source, modules, array operations, object orientation, and parallelism. Yet the standards remain largely backward-compatible, and much of the deployed code never moved forward with them. A great deal of it is still fixed-form FORTRAN 77, a style many scientific packages kept long after Fortran 90 made it optional. These legacy codes carry numerical methods that current science depends on while obstructing the tooling, threading, and accelerator work that future science requires. Rewriting them in a modern standard is conceptually unremarkable and practically unbounded: the transformations are individually routine, the codebases are enormous, and the work has therefore gone undone across much of computational science.

Modernizing legacy Fortran, then, is a problem of volume rather than difficulty, and it is exactly the kind of volume that agentic AI, capable frontier models driven through automated loops, can now absorb. An agent that reads a file, rewrites it, compiles it, runs a test, and iterates until the test passes drives down the per-file human cost that left the work undone. That makes agentic modernization an attractive answer, and it raises the central question of this paper: the extent to which such work can actually be delegated, and the factors that bound it. But converting tens of thousands of lines of dense integral code does not so much remove the bottleneck as relocate it: no one will read the machine-written result to check it by hand. The question is how anyone would know the rewrite was wrong. For a quantum-chemistry kernel, being wrong need not mean crashing. It can mean a silently different digit in the twelfth decimal place of a self-consistent-field energy, propagating into a published result years later. The limit on delegation, then, is set not by how much an agent can write but by how much can be checked. In the present setting that constraint is a benign one, because legacy scientific codes are unusually well supplied with exactly the kind of checking that is required.

The code we modernized is one such package. GAMESS [3–6], the General Atomic and Molecular Electronic Structure System, computes molecular electronic structure from first principles, covering Hartree-Fock, density-functional theory, perturbation theory, coupled-cluster theory and multireference methods, as well as fragment approaches including the effective fragment potential [7]. The development of GAMESS began on 1 October 1977 under the National Resource for Computation in Chemistry, supported by the National Science Foundation; the codebase split into US and UK branches in 1981, and the US branch has since been maintained by the Gordon group, first at North Dakota State University and since 1992 at Iowa State University. It exceeds a million lines of Fortran, the overwhelming majority fixed-form F77 written between 1985 and the present.

Two of its properties make GAMESS especially well suited to this study. The first is its distribution model: GAMESS is source-available freeware rather than open source, free to download under mild license restrictions. A code that cannot accept drive-by community contributions accumulates exactly the sort of maintenance debt of interest here, so the licensing is part of what makes it representative, though it also means we cannot ship the modernized source as an artifact. The second property is the one that makes this study possible. GAMESS provides forty-nine standard test inputs, and its user community treats the outputs as canonical reference values, reproduced to the last digit. We supplement these with two additional calculations to broaden the validation coverage. This convention is more than documentation: It supplies a ready-made, exact, community-sanctioned definition of correct behavior.

Within the GAMESS codebase, we selected the electron-repulsion-integral (ERI) machinery that evaluates four-center integrals, $(\mu\nu \mid \lambda\sigma)$, in shell quartets. The eight *int2* files span four ERI implementations: rotated-axis packages for (s,p,L) and (s,p,d,L) shell quartets that combine Pople–Hehre axis rotation with McMurchie–Davidson quadrature [8–10]; the Electron Repulsion Integral Calculator (ERIC) for selected (s)- through (g)-shell quartets [11]; and the general Rys-quadrature implementation [12–14]. Together with four related files: *eftei*, *eftei_eric*, *eftei_genr03*, and *eftei_genr70*, which we refer to collectively as the four-file ‘eftei‘ family these constitute the twelve-file conversion scope. Every SCF iteration, every direct-MP2 amplitude, every coupled-cluster contraction, every gradient and Hessian, and every coupling term to the effective fragment potential passes through the ERI section of the code. It is simultaneously the most valuable part of the package to modernize, since COMMON blocks and fixed-form layout there obstruct threading and GPU offload, and the most unforgiving, since its outputs are physical observables that everything downstream consumes. Modernizing peripheral utility code would have established little; we deliberately targeted the component a careful maintainer would be most reluctant to touch.

With the target chosen and an exact check for correctness in hand, we state the central claim of the paper: *a supervised agentic workflow can carry out legacy scientific-code modernization at production scale, with the specification rather than any model as its durable artifact; the reach of that delegation is set by the verification the domain affords, and scientific software affords more of it than is usually recognized.* We support this claim with four contributions. 1) We design the workflow itself: three prompt-specialized roles operating under a version-controlled specification that the agents wrote and revised. 2) We distinguish what conversion mechanically requires from what it requires comprehension to do, and we locate the agentic contribution in the latter. 3) We show verification performed largely *by* the agents rather than merely *on* them, and report the outcome across all twelve files under two independent checks: an exact, bit-for-bit battery against baselines the agents captured, and the maintainers’ own tolerance-based continuous-integration suite. 4) We characterize the resulting automation boundary empirically: what the agents did unaided, the one defect class that escaped them, the escape their verification could not structurally detect, and where human effort remained.

The remainder of the paper is organized as follows. Section II positions the work against modernization in comparable packages, rule-based Fortran tooling, and agentic code transformation. Section III describes the agentic workflow: its roles and specification, the demands a conversion places on the agents, and their self-verification. Section IV reports the results across all twelve source files. Section V discusses the workflow failures, the implications for the boundary of safe delegation, and the limits of the evidence. Section VI concludes the paper.

## II. Related Work

Three lines of research are relevant to this work. The first is the practical experience of modernizing large scientific packages, which shows the approaches comparable projects have taken to legacy Fortran and the cost of each. The second is the decades-old body of automated, rule-based Fortran restructuring tools that predate language models and already perform many of the mechanical transformations we report. The third is the recent and fast-growing literature on agentic code transformation, in which language models plan and iterate over code. We position our work against each in turn, and then, in a fourth subsection, set our zero-tolerance verification criterion against established practice, the point at which our approach most clearly departs from prior work.

### A. Modernization in Comparable Packages

Every large quantum chemistry package has faced the F77 question and answered it differently. Psi4 [15–17] was rewritten from Fortran into C++ in the early 2010s, a clean-slate effort that consumed multiple person-years and broke continuity with PSI3. NWChem [18] kept a hybrid F77/F90 surface and has migrated hot loops incrementally for over a decade. However, it is necessary to mention NWChemEx was developed in response to the Exascale Computing Project and was written specifically to take advantage of accelerators such as GPUs. CP2K [19] has been free-form Fortran with explicit modules almost from inception, and Quantum-ESPRESSO [20, 21] has progressively wrapped modules around plane-wave kernels while leaving legacy F77 in the lower-level recurrences.

The same incremental pattern recurs outside chemistry. ELPA [22], the eigensolver library several of the aforementioned packages depend on, moved to Fortran 2003 modules with explicit interfaces, and its maintainers report that compile-time interface checking exposed rank- and type-mismatch bugs that had survived in production, a finding our own compile-time results reproduce.

Our conversion envelope is closest to that of NWChem. The difference lies in the execution, which is agentic, and the acceptance criterion, which is zero-tolerance equivalence against the published tests of the package. We have not found that criterion documented in other efforts; tolerance bands are the norm.

### B. Rule-Based Fortran Restructuring

Automated F77 modernization long predates language models, and commercial restructuring tools have existed for decades: vendor documentation for plusFORT and its SPAG component [23] describes fixed-to-free-form conversion, `GO TO` restructuring into block constructs, declaration generation, and dead-code elimination, and comparable format converters are widely available. A substantial share of the edits we report is work such tools already handle. We have not run them against our matrix and make no measured comparison.

The distinction we claim is therefore not between manual and automated work, but between mechanical source-to-source transformations and tasks that require semantic or architectural comprehension. Rule-based transformers can perform the former, but they do not notice when a declaration contradicts its own usage, do not reason about circular dependencies across a file group, do not write a unit-test driver with domain-appropriate edge cases, and do not compose a new audit rule after diagnosing a defect they introduced. The present paper separates these mechanical and comprehension-dependent tasks explicitly, and shows a case where the difference decided the outcome.

### C. Agentic Code Transformation

A growing literature applies language models to code translation, refactoring, and test generation, and increasingly to agentic loops that plan and iterate over whole repositories [24]. Evaluation in this literature runs largely on benchmarks of small self-contained programs judged by functional correctness, of which SWE-bench [25] is representative. Closer to our target, a small but growing line of work applies LLMs and agentic loops to Fortran specifically: dual-agent Fortran-to-C++ translation with iterative compile–test–fix loops [26], a cross-platform LLM-assisted Fortran-to-C++ study [27], and an autonomous agentic workflow porting legacy Fortran to portable Kokkos [28]. This line of work, ours included, rests on two recent advances: frontier models capable enough to comprehend and rewrite production code, and general-purpose agent runtimes that supply the plan–compile–test–iterate loop.

Our setting differs from all of the aforementioned work, including the Fortran-specific efforts, on three axes we think are underexplored. First, the transformation is in-language and semantics-preserving: it is F77 to F2008, not a port to C++ or Kokkos, so the original numerical behavior must be retained exactly rather than re-expressed. Second, the acceptance criterion is exact numerical reproduction rather than the functional or unit-test correctness these efforts adopt, which matters because a plausible-looking wrong answer can pass a test suite but cannot reproduce a twelve-digit energy. Third, the unit of work is one file inside a million-line call graph whose other callers cannot be modified alongside it (cf. also Kokkos [28]).

### D. Reproducibility as an Acceptance Criterion

Bit-for-bit reproduction rests on IEEE-754 determinism together with the absence of floating-point-reorganizing transformations; `-O1` compilation and a conversion rule set that forbids algorithmic change satisfy both. The approach resembles golden-output testing in compiler-validation suites, except that our reference standards (denoted *goldens*) are physical observables from production-equivalent runs rather than outputs of synthetic programs. Trilinos [29] offers the closest HPC precedent, differing against retained historical outputs under tolerance. We set the tolerance to zero for the standard tests provided by GAMESS, which makes the verdict mechanically decidable, a property whose value rises sharply when the author under review is an agent and the reviewer cannot read the difference.

## III. The Agentic Workflow

This section describes the agentic workflow in three parts: the setup and governance of the agents, the requirements of file conversion, and output verification.

### A. Roles, Specification, and Gates

The workflow is the part of this system that transfers. It consists of three prompt-specialized agent roles, a version-controlled specification that governs them, and a small number of human gates. Humans write no code and run no conversions; they hold the decisions where automation should not act alone: approving side-effecting shell commands, reconciling build configuration, adjudicating whether results of a converted file are acceptable, and approving each merge. The agents authored the specification and executed it, but not without oversight, since every change to the specification, and every converted file, reaches the shared branch only through a human-approved commit. Its unusual property is that the rules were agent-written and agent-revised as the work exposed new cases, while the judgment of when a result, or a new rule, is good enough stayed with the humans.

*1) Defining the Agent:* We use *agent* to mean an LLM-backed process that pursues a stated goal across multiple steps without per-step human direction, using tools that modify persistent state, e.g., reading and writing files, invoking a compiler, running tests, operating version control, under a specification loaded into its context when the session begins.

Concretely, the agents here are sessions of Claude Code, the command-line coding agent of Anthropic, which supplies the loop, the tool surface (file read/write, shell execution, version control), the permission prompts that gate side effects, and the convention by which a `CLAUDE.md` file at the repository root is loaded into context automatically at the start of every session and points the agent to the rest of the specification described below. That off-the-shelf machinery is load-bearing, not incidental: the workflow builds on it rather than reimplementing it. Four model generations ran the work: Claude Opus 4.6, Claude Sonnet 4.6, Claude Opus 4.7, and Claude Opus 4.8. We name the harness and the models because portability across them is one of our findings, not because the approach is model-independent: it does depend on a capable frontier model and a general agent runtime. It does not, however, depend on any particular Claude generation. We have not tested other model families, and treat comparable-capability portability as a reasonable conjecture rather than a result.

This is not a multi-agent system in the coordination sense. The roles do not negotiate, share runtime state, or run against one another; they are prompt-specializations of the same tool, distinguished by the instructions loaded into context rather than by being separate processes. The arrangement costs autonomy but buys reproducibility, auditability, and model-portability. One further limit is worth stating: the per-file loop is prescribed by the specification rather than planned per file. Agents plan the conversion strategy inside it: which constructs to restructure, which control flow is irreducible, how to sequence declaration inference across twelve thousand lines.

*2) Three Roles:* The conversion role reads the original `.src` file, writes the modernized `.f90`, the change log, and the unit-test driver, captures the golden output from the F77 build, and verifies the modernized build against it.

The testing role checks the branch out into its own worktree, compiles both f77 and f90 binaries, compares their results, and designs additional numerical tests. The specification names the cases for which an integral kernel is most prone to precision loss such as very tight and very diffuse Gaussian exponents. It then posts a structured pass/fail report to the pull request.

TABLE I
STYLE-COMPLIANCE CHECKS EXECUTED BY THE REVIEW-ROLE AGENT.

| Check | Rule |
|---|---|
| `DOUBLE PRECISION` in code | must be absent |
| Old relational operators | `.EQ. .NE. .LT. .GT. .LE. .GE.` absent |
| `GO TO` in code | flagged; retained instances must be annotated |
| Line length | <= 132 columns |
| `IMPLICIT NONE` count | must equal number of program units |
| `REAL(REAL64)` uppercase | must be zero |
| `USE ISO_FORTRAN_ENV` uppercase | must be zero |
| COMMON `TODO` markers | must equal COMMON statement count |
| Module structure | one `MODULE`, `PRIVATE` default |

The review role audits style compliance through the checks in Table I and posts the results as a table. Most of these are unremarkable greps, but the COMMON block check is not: requiring the count of deferral annotations to equal the count of COMMON statements makes silently skipping deferred work mechanically detectable, which is the sort of check that matters precisely because nobody would notice its absence.

Each of these roles works in its own git worktree, so they can run without interfering with each other, whether on a single machine or across separate machines run by different contributors. Each sees exactly the state of its branch, and they coordinate only through the shared repository. The separation is deliberate: the testing and review roles exist to check the work of the conversion role independently, and that check is only as trustworthy as it is independent. Ideally, a different contributor drives each role; a single operator can run all three, but then the judgment that produced a file also clears it, weakening the independence the split exists to preserve.

*3) The Agent-Authored Specification:* Three documents govern the conversion. `CLAUDE.md` defines the workflow contract and is loaded automatically into every session. `STYLE_GUIDE.md` defines the conversion envelope as rules. `AGENT_GUIDE.md` holds the role prompts, compiler invariants, and limitations.

All three are agent-authored, as are their revisions. This is not incidental. The specification is not scaffolding wrapped around the agents by their operators; it is the workflow itself, written by the system that then ran it. Two revisions make the point concretely. In the first, an agent restructured its own context file, moving role prompts, compiler flags, and worktree instructions out of `CLAUDE.md` and into `AGENT_GUIDE.md`, shrinking the auto-loaded file from 243 lines to 176. The stated reason was to reduce the context loaded every session. An agent trimming its own operating budget is difficult to describe as anything but agentic work. The second revision followed a defect an agent had itself introduced: after diagnosing it, the agent wrote a new audit rule into `STYLE_GUIDE.md` so the whole defect class could not recur (see Section V-A for details).

*4) The Conversion Loop and Its Artifacts:* Each file yields exactly four committed artifacts: the modernized source, a change log documenting every category of change, a unit-test driver, and the golden output captured from the F77 build.

The conversion loop runs as follows: the agent branches from `master`; reads the original; writes the modernized source, the change log, and the driver; compiles the module dependencies in fixed order; compiles the original in fixed form and captures the golden; compiles the modernized source under `-std=f2008` and diffs it against that golden; builds a full GAMESS binary with the file substituted and runs the 51 standard tests and the Jenkins tests; and finally submits for human review.

The four-artifact contract carries more weight than its simplicity suggests. It makes each file evidence self-contained, so a reviewer can audit one file without reading the other eleven. A rule-based transformer emits transformed source, not the evidence to check it (i.e., the unit-test driver, the golden, and the change log) and requiring the agent to produce that evidence keeps per-file review tractable.

One implementation detail explains why per-file conversion is possible in a codebase nobody can convert atomically. Wrapping a file in a `MODULE` breaks the flat symbol references that thousands of unmodernized F77 callers depend on, so a clause in the build driver strips the module wrapper at compile time. Ten of twelve files need only this. Two keep their `USE` statements at module scope, where subroutines inherit imports by host association that evaporates when the wrapper is removed, and these require the imports to be injected per subroutine instead.

*5) Human Oversight:* The specification states the limits explicitly: agents cannot approve or merge pull requests, they post review comments rather than formal approvals, and shell commands require interactive approval, so an operator must be present. Merging additionally requires two human approvals and a passing CI run. We characterize this as supervised rather than autonomous by design.

### B. Conversion Requirements

If conversion is mechanical, agentic execution is an expensive way to do it and, as we noted earlier, rule-based tools have handled much of it for decades. This section separates the part that is mechanical from the part that is not, because the distinction determines where the useful agentic contribution actually lies, and where the failures land.

*1) The Envelope:* The style guide fixes the target against the Fortran 2008 standard [30]: free-form 132-column source, `IMPLICIT NONE` everywhere, `REAL(real64)` from `iso_fortran_env`, `INTENT` on all dummy arguments, structured control flow in place of `GO TO` and arithmetic `IF`, modern relational operators, character literals for Hollerith constants, one module per file, and COMMON blocks preserved verbatim under an explicit deferral annotation.

Two exclusions are deliberate. COMMON migration is unsafe file-by-file, because the blocks are shared across many files. Algorithmic refactoring, such as floating-point reorganization, loop fusion, and vectorization, is forbidden outright, and this second exclusion is load-bearing in a way that is easy to miss. Permitting it would forfeit exact equivalence and force a retreat to tolerance bands, which would reintroduce reviewer judgment at precisely the scale where reviewer judgment fails. The envelope is chosen to preserve the oracle.

Modernization removes the structural barriers to threading and offloading, such as fixed-form layout, COMMON blocks, and unstructured control flow, while performing none of the arithmetic reordering those optimizations require; that reordering belongs to a later phase under a tolerance-based criterion. Certifying under bit-for-bit equivalence that the conversion changed nothing is the precondition for optimizing on top of it, not a contradiction of it.

*2) The Mechanical Majority:* Based on line volume, conversion is dominated by transformations a rule-based tool can perform: fixed-to-free form, `.EQ.` to `==`, `C` to `!` comments, `1.0D+00` to `1.0_real64`, `DOUBLE PRECISION` to `REAL(real64)`, `PARAMETER` restyling, and most numbered `DO` loops. We do not claim these as an achievement.

*3) The Comprehension Minority:* By difficulty and by risk, conversion is dominated by a much smaller set of decisions that require reading the file as a whole.

Satisfying `IMPLICIT NONE` means inferring type, rank, and dimension for every undeclared variable from how it is used, then confirming that usage does not contradict the implicit default it previously relied on. Annotating `INTENT` means classifying every dummy argument as read, written, or both. Restructuring control flow means determining which label-jump graphs are reducible, and, just as importantly, which are not.

Some decisions require reasoning beyond the file. In the four-file eftei family defined above, the dispatcher calls into its three companions; placing `USE` statements at module scope would have created circular dependencies, so the agent placed them inside the calling subroutine instead. That is architectural reasoning about a file set, arrived at without being told the cycle existed. Others require knowing why the original code is written the way it is.

*4) Judgment:* Three cases illustrate the difference between following rules and exercising judgment.

The agents retained 718 `GO TO` statements (679 in one file, 37 in another and 2 in a third) inside Rys quadrature drivers and computed-`GO TO` dispatch tables that cannot be restructured without altering dispatch semantics. Each retention carries an explicit annotation. An agent optimizing for apparent style compliance would have restructured them and risked the arithmetic; these were left deliberately and marked. Knowing that 718 annotated

```
PROGRAM TEST_INT2D
  USE, INTRINSIC :: iso_fortran_env, &
       ONLY: real64, int64
#ifdef USE_MODULE
  USE INT2D_MOD
#endif
  IMPLICIT NONE
  INTEGER, PARAMETER :: NTRIALS = 10000
  INTEGER :: ISEED, ITRIAL
  REAL(real64) :: RESULT(4)
  ISEED = 17  ! deterministic seed
  CALL RANDOM_SEED(ISEED)
  DO ITRIAL = 1, NTRIALS
    CALL DSPS(...)  ! fills RESULT
    WRITE(*, '(I8, 4F22.15)') ITRIAL, RESULT(1:4)
  END DO
END PROGRAM
```

Fig. 1. A synthetic unit-test driver. The same source is compiled once against the F77 original and once against the modernized module via `-DUSE_MODULE`, and the outputs are diffed.

items remain in three named files is a considerably better position than believing there are none.

Forced explicitness surfaced defects that predated the modernization. One routine declares `COMMON /SHLINF/` with two integers while another declares the same block with four. They access disjoint portions, so the inconsistency had survived for years behind a compiler warning. Several call sites pass scalars where rank-2 arrays are expected. The agents documented both as known issues in the original. Each change log also carries a deferred-items section naming every retained COMMON block and every unresolved external reference. Additionally, the agent records what it chose not to do, which is the part of a conversion that usually goes unwritten.

### C. Verification, Performed by Agents

Verification is a step inside the workflow rather than a separate activity, and agents perform most of it. They wrote the test drivers and chose the edge cases, captured the golden baseline reference from the F77 build, built the bisection harness that later localized the one real defect, ran the style audits, and audited their own output against the original. This section describes the two layers and the criterion that decides pass from fail.

*1) Synthetic Unit Tests:* Each file ships a driver, written by the conversion agent, that is compiled twice from identical source (once against the F77 original, once against the modernized module) with the outputs diffed. Figure 1 shows one such example.

A fixed seed and a fixed-width output format with no timestamps make the comparison deterministic. Roughly 255,000 synthetic shell quartets are exercised across the matrix, with no differences.

The limitation of the layer is that synthetic quartets do not reproduce the dispatch sequence a real SCF run generates, and in particular never reach paths gated by buffer-overflow logic under sustained enumeration. The one defect that escaped conversion lived exactly there.

*2) Integration Testing:* The second layer rebuilds a complete GAMESS binary in which exactly one file differs, then validates that binary in two independent ways. One compares its output bit-for-bit against results generated from the F77 build, across the package-published test inputs: all forty-nine standard exams, spanning the method families, run types, and basis sets the package supports; a cc-pVQZ water calculation that is the only member reaching angular momentum beyond (L=2); and a scalar-relativistic DFT calculation on HCl that is the most expensive single test at about fifteen minutes. The other uses the maintainers' own Jenkins continuous-integration suite, which runs the project regression tests and compares each result against the expected values GAMESS ships, using a tolerance-based checker rather than exact reproduction.

Because a raw log contains hundreds of run-varying lines, the bit-for-bit check filters each log to its chemistry-relevant content: energies, iteration traces, nuclear repulsion, virial ratios, gradients, and convergence markers, discarding timestamps, hostnames, and wall-clock values. A classifier then labels each test PASS when nothing chemistry-relevant differs from the reference results, TIMING when only wall-clock lines differ, and DIFF when any chemistry-relevant line does. DIFF is failure, and merging is gated on every test being PASS or TIMING.

*3) Agent Self-Audit:* An audit clause requires each file to be checked for `ISHFT`-based packed-label construction, and agents execute it as a differential count against the original. In the four-file eftei family, for instance: the modernized code matched the original on every count: six bit-manipulation calls, seven packed-label array parameters, and six subroutines writing those labels, with none of the label arrays declared as a real type.

An agent verifying its own conversion by counting its output against the input, and committing the comparison for a reviewer to check, is a modest mechanism. Combined with the review role auditing the work of the conversion work, it means agents perform verification on agent output while humans hold the merge gate.

## IV. Results

This section reports the outcome across the twelve files: the scope of the conversion, the verdict on every file, and the division of labor between agents and humans. These results set up the defect analysis that follows.

### A. Scope

Table II reports the conversion scope, including the old F77 line count, the new F90 line count, percentage of change of line count, the number of subroutines, the number of GO TO statements, and the number of COMMON blocks for each file, from Column 3 to Column 8.

TABLE II
Conversion scope and test results.

| Family | File | .src | .f90 | Delta | Subs | GO TO | COMMON | Standard Tests | Jenkins Tests |
|---|---|---|---|---|---|---|---|---|---|
| int2 | int2a | 12,238 | 11,498 | -6.0% | 38 | 679 | 361 | Pass | Pass |
| int2 | int2b | 2,975 | 2,629 | -11.6% | 19 | 0 | 77 | Pass | Pass |
| int2 | int2c | 4,685 | 4,894 | +4.5% | 22 | 0 | 52 | Pass | Pass |
| int2 | int2d | 5,793 | 6,402 | +10.5% | 21 | 0 | 40 | Pass | Pass |
| int2 | int2f | 4,428 | 4,635 | +4.7% | 15 | 0 | 28 | Pass | Pass |
| int2 | int2g | 1,820 | 1,890 | +3.8% | 6 | 0 | 12 | Pass | Pass |
| int2 | int2r | 9,060 | 8,958 | -1.1% | 37 | 2 | 172 | Pass | Pass |
| int2 | int2s | 6,835 | 6,997 | +2.4% | 17 | 0 | 15 | Pass | Pass |
| eftei | eftei | 4,071 | 3,849 | -5.4% | 16 | 37 | 122 | Pass | Pass |
| eftei | eftei_eric | 2,093 | 1,200 | -42.7% | 12 | 0 | 48 | Pass | Pass |
| eftei | eftei_genr03 | 1,371 | 1,528 | +11.5% | 11 | 0 | 77 | Pass | Pass |
| eftei | eftei_genr70 | 1,079 | 1,436 | +33.1% | 11 | 0 | 62 | Pass | Pass |
| **Total** | | **56,448** | **55,916** | **-0.94%** | **225** | **718** | **1,066** | | |

Aggregate line count is essentially unchanged, at -0.94% of the original F77 line count. Individual files range from -42.7%, where inline-expanded angular-momentum cases collapsed into shared kernels, to +33.1%, where COMMON indirections became explicit arguments. The flatness in aggregate is mildly interesting as evidence against a failure mode one might reasonably expect from LLM-authored conversion: the agents neither padded nor over-compressed at scale.

Subroutine count is preserved exactly in every file. Nothing was added, removed, split, or merged. This specification requirement was held because it is mechanically checkable. All 1,066 COMMON statements survive verbatim with their deferral annotations.

### B. Outcomes

Every modernized file was validated two ways. The battery compares, for each binary, chemistry-relevant output bit-for-bit against a golden reference captured from the Fortran 77 original, where a difference in the twelfth decimal place counts as failure; each binary runs the full 51 standard tests. The Jenkins set instead validates against the expected values GAMESS ships with its own suite, using the maintainers' tolerance-based checker, and adds 49 tests per file. Every file passes both checks, with no chemistry-relevant difference in any run; the only lines that ever differ are wall-clock timings. The last two columns of Table II show the test results. In addition, the union build of all 12 modernized files passed the standard tests and Jenkins tests as well. For artifact description and evaluation, we refer the reader to two GitHub repositories [1,2].

### C. Effort and Division of Labor

The twelve-file validation sweep ran sequentially in a single session totalling 7 hours 29 minutes, which is about 37 minutes per file. Since the standard tests take about twenty-five minutes, roughly a third of the time went to building, staging, adjudication, and documentation for each file, and thereby the time grows only with the number of files rather than with their size.

Eleven of twelve files were arithmetically correct on the first conversion, and eleven of twelve compiled cleanly on the first attempt. Nine of twelve had valid unit-test goldens on the first capture, and the three exceptions are the subject of Section V-A.

[1] https://github.com/gms-bbg/test-validation

[2] https://github.com/shenyuzhong/gamess-modernization-paper

## V. Discussion

### A. Defect Analysis

Three classes of defect arose during the conversion. Each was caught by a different mechanism, and one by no automated mechanism, so we treat them separately below. Nine of the twelve files required no post-conversion correction.

*1) Defects Caught by the Compiler:* Strict `-std=f2008` is a considerably harsher regime than the legacy build, and it caught two classes during the conversion of one file. Scalars passed where rank-1 dummy arguments are declared, tolerated for decades under `-std=legacy`, became hard errors and were fixed with array constructors at the call site. Hollerith literals in `DATA` and `FORMAT` statements were rejected outright and rewritten.

Neither affected runtime arithmetic, and both were resolved before any integration test ran. Tightening the language standard is a cheap first line of defense for agentic conversion, and it catches a distinct class, namely, interface and literal-form errors, at no marginal cost per file.

*2) The Escaped Defect:* One class survived both the agents and the compiler. GAMESS packs four shell indices into a single 64-bit word, reusing the same eight-byte slot for packed labels and for integral values, and writes the bit pattern through a real-typed alias:

```
DOUBLE PRECISION IX(*)
IX(K) = ISHFT(I,48) + ISHFT(J,32) + ISHFT(K,16)
+ L
```

The agent read that declaration and translated it faithfully as `REAL(real64) :: IX(*)`. The result compiles cleanly, satisfies every clause of the style guide, and is a locally accurate rendering of what the original says. It is also wrong, because the high-order shifts no longer land where they must.

What makes this characteristically agentic is where the error sits. The original declaration had always contradicted the code semantics, so reproducing it faithfully is precisely what careful local translation produces. Local fidelity, usually an agent's virtue, is the bug here; a rule-based transformer would have made the identical mistake for the identical reason. That is the clearest evidence that this part of conversion is not mechanical.

The unit tests never saw it. The affected path is gated by buffer-overflow logic that only fires under sustained shell-quartet enumeration, and ten thousand synthetic trials never got there. The standard tests caught it on the first run of `exam01`: the first SCF iteration came back at -61.29 Ha where the baseline said -37.17, the modernization then froze at -68.157, and the run terminated with a failure to locate a stationary point.

*3) Self-Repair by the Workflow:* An agent localized the regression by subroutine-level bisection, recompiling repeatedly with one subroutine at a time reverted to its original, and narrowed 38 subroutines to the culprit in five rounds. The bisection harness did not exist beforehand; the agent wrote it during the investigation (113 lines of Python) and retained it as tooling. The correction itself was a single declaration: `IX` retyped from `REAL(real64)` to `INTEGER(int64)`, so the byte shifts compose a packed integer again.

The agent then generalized from the instance to the class (any `ISHFT`-based packed-label construction where the declared type of the recipient is too narrow or wrongly typed) and wrote that generalization into `STYLE_GUIDE.md` as an audit clause requiring every subsequent file to be checked for it. Because the specification loads into every session, this changed the behavior of every conversion that followed. Applied retrospectively, the clause found two more instances in the file just fixed and exactly one other (clean) file that used the construct. The total corrective change was six wrong lines out of 55,916, albeit having even one wrong line is unacceptable.

We offer that ratio cautiously, since one project cannot establish an error rate. The narrower suggestion is more actionable: defects that survive a strict compiler here appear few, tightly clustered, and semantically deep. The profile that defeats sampling-based review and rewards exhaustive exact checking. Six lines will not be found by reading 55,916; they are found immediately by running forty-nine tests.

*4) A Common-Mode Verification Blind Spot:* The unit-test drivers for three converted files had been captured before `-fdefault-integer-8` was standardized across the build. Regenerated under the standard configuration, they differed substantially, in 26, 36, and 105 changed line pairs.

While the flaw persisted, no differential check could have caught it. Both sides of the comparison, the golden and the binary checked against it, were built with the same missing flag, so the diff was empty and the test passed: internally consistent, externally invalid. What surfaced was not the check but the build being standardized; once the correct flag reached the binary while the stale unit test driver still carried the wrong one, the comparison showed the difference.

The general form constrains any verification strategy built on differential comparison: *a differential check cannot detect an error in the procedure that generates both of its sides.* It is blind along its common mode, and the error surfaces only when something outside that procedure, such as a later standardization of build flags, breaks the symmetry. The standard tests avoid the trap because their results were generated once from a reference production build rather than regenerated alongside the artifact under test. The Jenkins tests provide a second, externally anchored reference that shares no common mode with those reference results, since they check against the expected values GAMESS provides independently. Together they support anchoring at least one verification layer to an externally fixed reference.

*5) Defects in the Reference Oracle:* Three defects turned up in the verification tooling itself: a timing-classifier regex missed one summary-line format and misreported clean runs as failures; the GAMESS driver script used a csh construct that silently misparsed argument counts; and an input-staging path failed for the one test outside the standard directory. The first and third should worry a reader, because in both the oracle returns the wrong verdict. An exact criterion is only as trustworthy as the harness computing it, and one that produces false failures teaches reviewers to discount failures.

### B. Locating the Automation Boundary

The case study generalizes beyond GAMESS along three dimensions: the automation boundary lies further out than caution would suggest; its shape matters more than its position; and the safety of delegation reflects a property of the domain rather than of the agents.

*1) Delegable Work:* Table III summarizes the division. Everything in the left column was performed by agents without human authorship; everything in the right column was not. The striking entries are at the bottom left: diagnosing a regression by bisection and writing a rule against the defect class into the governing specification (Section V-A3) are not tasks we expected to delegate.

*2) The Role of Code Structure:* Human effort did not scale with code volume. Nobody reviewed 55,916 lines, and no amount of additional code would have changed that. Instead the human effort concentrated at points where the automated checker's own assumptions could fail, e.g., in configuration reconciliation, adjudication, and the merge decision.

This is a considerably more scalable place to spend attention than reading output. It also predicts where to look when something goes wrong: the one escape that no automated check could catch was found by a human doing exactly the kind of cross-checking that the automated layer structurally cannot perform on itself. This generalizes better than any specific number here. The useful question when planning agentic modernization is not *how much can be automated* but *where can automated checking be wrong about itself*, and that is where the humans belong.

TABLE III
THE OBSERVED AUTOMATION BOUNDARY.

| Delegated to agents | Retained by humans |
|---|---|
| Reading and comprehending 56,448 lines of F77 | Approving side-effecting shell commands |
| Emitting conforming Fortran 2008 | Build-driver and baseline plumbing |
| Authoring unit-test drivers and edge cases | Branch merges |
| Capturing golden references | Adjudicating whether results are acceptable |
| Documenting changes, deferrals, known issues | Reconciling build configuration |
| Running style-compliance audits | Approving pull requests |
| Executing the integration battery | |
| Diagnosing a regression by bisection | |
| Writing new specification rules from a post-mortem | |

*3) The Specification as Durable Artifact:* The competence the system accumulated was kept in version-controlled text by design, rather than left to any model or session memory. The work spanned four Claude model generations (Opus 4.6, Sonnet 4.6, Opus 4.7, and Opus 4.8) and no rework was required when the underlying model changed because that competence lived in the specification: the four-artifact contract, the compiler-flag discipline, the audit clauses, and the style checks. For scientific software, where provenance is the point, an adaptation mechanism that can be diffed, reviewed, and reverted seems preferable to one that cannot.

### C. Limitations

This work has several limitations. First, all the systematic tests described in this paper were carried out on a desktop workstation, and they have yet to be extended to HPC clusters. We also validated the results against only the standard tests and the Jenkins test set, but not the extended test suites that GAMESS provides, so code paths reached only by those additional tests remain unchecked on our side. Validation against the extended GAMESS test suite is underway and will be reported in more detail in a subsequent paper. Moreover, none of these changes have yet been integrated into the production version of GAMESS. Second, we used only Anthropic tools, namely the Claude Code runtime and four generations of Claude models. The workflow was not exercised on other agent runtimes or model families, and we treat comparable performance from models of similar capability as a conjecture rather than a demonstrated result. Third, the study covered a single codebase. GAMESS is unusually well suited to this approach because its test suite provides an exact check on correctness; codebases that lack such a check may not support the same workflow.

## VI. CONCLUSION

This study evaluated whether a supervised agentic workflow could sustain legacy scientific-code modernization at production scale. The two-electron-integral core of GAMESS (twelve files, 56,448 lines, 225 subroutines) provided a demanding test, and the workflow converted it from Fortran 77 to Fortran 2008 almost entirely through agents, with every file reproducing the test outputs published to the last printed digit and passing the continuous integration test suite.

The contribution of this study lies less in the modernized code than in an account of how such work can be delegated, and how far. The workflow itself, three prompt-specialized roles under a version-controlled specification that the agents wrote and revised, keeps the project-accumulated competence in that specification rather than in any model. Within a conversion, the mechanical transformations separate cleanly from the parts that require comprehension, and the agents earn their place in the latter, since a rule-based transformer fails at exactly those points. Verification, notably, was carried out largely by the agents rather than performed on them, and the resulting boundary can be read from the record: the work the agents completed unaided, the one defect class that slipped past them, the escape their differential check could not have caught, and the handful of points at which human judgment stayed essential.

Running beneath all of this was a single enabling property: a strict check, inherited from a domain that treats its published energies as canonical. That check, more than the agents' capability, set how far delegation could safely reach, and legacy scientific software supplies such checks more readily than is often assumed. Carrying the approach further is largely a matter of scale and coverage: integrating the modernized files together rather than one at a time, running the extended test suites GAMESS provides, extending the workflow across the rest of the roughly million-line package, and eventually converting the COMMON blocks into modules that a single-file change cannot reach, toward a fully modernized GAMESS.

## ACKNOWLEDGMENTS

This work was supported in part by the U.S. Department of Education under the grant # P116S230016 and in part by a grant from the Air Force Office of Scientific Research, under contract AFOSR FA9550-18-1-0321.